\documentclass{article}
\usepackage{spconf,amsmath,graphicx}
\usepackage{epstopdf}
\usepackage{array}
\usepackage{graphicx}
\usepackage{multirow}
\usepackage{color, soul}
\usepackage{cite}
\usepackage{url}
\usepackage{hyperref}
\usepackage{booktabs}
\usepackage{amsmath}
\usepackage{subfigure}

\usepackage{amssymb}

\title{Emotion-Aware Prosodic Phrasing for Expressive Text-to-Speech
}
\name{Rui Liu $^{ 1}$\thanks{The research by Rui Liu was funded by the High-level Talents Introduction Project of Inner Mongolia University (No. 10000-22311201), the Young Scientists Fund of the National Natural Science Foundation of China (No.\ 62206136) and the Guangdong Provincial Key Laboratory of Human Digital Twin (No.\ 2022B1212010004). Haizhou Li is partly supported by National Natural Science Foundation of China (No. 62271432) and AME Programmatic Funding Scheme (Project No. A18A2b0046).}, Bin Liu $^{ 1}$, Haizhou Li $^{ 2,3}$}
\address{ $^1$ 
Inner Mongolia University, Hohhot, China \\$^2$ Shenzhen Research Institute of Big Data, School of Data Science, \\The Chinese University of Hong Kong, Shenzhen, China \\ $^3$ National University of Singapore, Singapore\\
\small{liurui\_imu@163.com, iframe\_liu@163.com, haizhouli@cuhk.edu.cn}
}
\begin{document}
%
\maketitle
\begin{abstract}
Prosodic phrasing is crucial to the naturalness and intelligibility of end-to-end Text-to-Speech (TTS). There exist both linguistic and emotional prosody in natural speech. As the study of prosodic phrasing has been linguistically motivated,  prosodic phrasing for expressive emotion rendering has not been well studied. In this paper, we propose an emotion-aware prosodic phrasing model, termed \textit{EmoPP}, to mine the emotional cues of utterance accurately and predict appropriate phrase breaks. We first conduct objective observations on the ESD dataset to validate the strong correlation between emotion and prosodic phrasing. Then the objective and subjective evaluations show that the EmoPP outperforms all baselines and achieves remarkable performance in terms of emotion expressiveness. The audio samples and the code are available at \url{https://github.com/AI-S2-Lab/EmoPP}.
\end{abstract}
%

\section{Introduction}
\label{sec:intro}  
 



 


Prosodic phrasing aims to break a long utterance into prosodic units using phrase break prediction \cite{vadapalli2016investigation}. Over the last few years, the Text-to-Speech (TTS) models have made significant improvement \cite{li2019neural,shen2018natural,ping2017deep} with the help of end-to-end architecture. Note that prosodic phrasing is often the first step in generating a prosody pattern, such as intonation and duration modeling \cite{liu2020exploiting}. Any errors made in the prosodic phrasing are propagated to the downstream prosodic models, resulting in unnatural speech. Therefore, prosodic phrasing is critical in improving the naturalness and intelligibility of TTS systems \cite{frazier2006prosodic}.

Traditional prosodic phrasing approaches mainly focus on the following two categories: 1) rich linguistic feature extraction and 2) effective architecture design. For the first category, some works attempted to incorporate the part of speech (POS) \cite{keri2007pause}, semantic and syntactic structure \cite{klimkov2018phrase}, contextual information \cite{watts2011unsupervised}, various high-level embedding representations \cite{si2022modelling}, and even multi-modal knowledge \cite{yi2023adversarial} et al. as the enriched input feature. For the second category, researchers tried to build the prosodic phrasing model with conditional random fields (CRF) \cite{qian2010automatic}, deep neural networks (DNNs) \cite{liu2018mongolian}, recurrent neural networks (RNNs) \cite{chen1998rnn}, bidirectional long short-term memory (BiLSTM) \cite{liu2018improving}, and self-attention-based transformer network \cite{du2019prosodic}, that allows for learning the long-term time dependencies and sequential characteristics in text. The above work has contributed greatly to improving the naturalness and intelligibility of TTS.

However, the influence of prosodic phrasing on expressive modeling, especially emotion, in TTS has not received much attention. Actually, different phrase breaks in the same utterance will express different emotions, and in turn, different emotional states will affect the placement of phrase breaks in an utterance \cite{duez1982silent}. When expressing nervous or anxious emotions, people may add more breaks to the sentence, making the voice more rhythmic and compact. There may be fewer breaks when expressing relaxed or composed emotions. Therefore, investigating the relationship between emotion and prosodic phrasing and incorporating this knowledge into the expressive TTS system to enhance its emotional expression will be the focus of this paper.

In this paper, we propose an emotion-aware prosodic phrasing model, termed \textit{EmoPP}, to contribute to the expressive modeling of TTS. Specifically, EmoPP consists of text encoder, emotion predictor, and decoder. The text encoder and emotion predictor aim to extract the linguistic feature and the emotion state from the input utterance, respectively. The decoder takes both the linguistic feature and the emotion state to predict the final phrase breaks relating to the emotional speech. In this way, EmoPP mines the emotional cues of utterance accurately and predicts appropriate phrase breaks.
The objective experimental results on IEMOCAP dataset suggest that our EmoPP outperforms all baselines in terms of break prediction accuracy. The subjective listening experiments with an expressive TTS model further validate our method.

The main contributions of this work can be summarized as follows:
1) We propose a novel emotion-aware prosodic phrasing model \textit{EmoPP} for expressive TTS; 2) We incorporate emotion information into the phrase break prediction model to learn the relationship between emotion and prosodic phrasing; 3) The objective and subjective experimental results validated our EmoPP. 
To our knowledge, this is the first emotion-aware prosodic phrasing scheme for expressive modeling of TTS.

\section{Emotion-Specific Prosodic Phrase Breaks}
\label{sec:dataobs}

We first conduct objective observations on ESD dataset \cite{zhou2022emotional} to validate the strong correlation between emotion and prosodic phrasing. For ESD, we note that each text was read aloud with five emotion categories (neutral, happy, angry, sad, and surprise), allowing us to easily compare the prosodic phrasing differences of the same utterance under different emotional states.

The original ESD dataset consists of 29 hours of recordings by 10 native English speakers and 10 native Chinese speakers. 
We just select the English subset, including 350 text transcriptions and 17500 audio recordings, to extract the phrase breaks in utterances for all audios. Specifically, we construct an automatic break extraction pipeline that includes \textit{Force Alignment} and \textit{Break Label Generation}. \textit{Force Alignment} employ Montreal Forced Aligner (MFA) \footnote{\url{https://montreal-forced-aligner.readthedocs.io/en/latest/}} to align the audio signal and word sequence. In \textit{Break Label Generation}, following \cite{yang2023duration} and according to the MFA results, we mark a word as ``1'' (means break) if it is followed by a silence segment of more than 30 milliseconds, otherwise it is ``0'' (means non-break).

Let $P^{n}$, $P^{h}$, $P^{a}$, $P^{sa}$, $P^{su}$ denote the prosodic phrase break sequence of five emotion states for the same sentence from one speaker. To make a comprehensive comparison of the similarities and differences in the prosodic phrasing of different emotions, we sample two emotion categories, $i$ and $j$, of five emotional states and calculate the Simple Matching Coefficient (SMC) \cite{smc}, denoted as $\xi_{i,j}$, between the phrase break sequence pair from two emotions.
\begin{equation}
    \xi_{i,j} = \frac{\sum_{s\in[1,M]}^{M}\sum_{t\in[1,N]}^{N}S\!M\!C(P^{i}_{t,s},P^{j}_{t,s} ) }{N*M}  
\end{equation}
where $N$ and $M$ mean the utterance and speaker numbers of ESD respectively.

\begin{table}[ht]
\caption{Simple Matching Coefficient (SMC) between prosodic phrase breaks of different emotions.}
\label{TABLE:1}
\centering
\small
\begin{tabular}{cccccc}
\hline
 & \multicolumn{1}{l}{Angry} & \multicolumn{1}{l}{Happy} & Neutral & Sad  & Surprise \\ \hline
Angry    & 1.00        & 0.91      & 0.92    & 0.92 & 0.91     \\  
Happy    &    -    & 1.00     & 0.91    & 0.90 & 0.90     \\  
Neutral  &   -     &   -      & 1.00    & 0.91 & 0.91     \\  
Sad      &    -     &     -     &  - & 1.00 & 0.90     \\  
Surprise &    -   &   -    & -& - & 1.00     \\ \hline
\end{tabular}
 \vspace{-5mm}
\end{table}


Note that the speaker-specific prosodic phrasing \cite{louw2016speaker,prahallad2010learning} is not our focus in this work. Therefore, we finally average the SMC scores from different texts and different speakers, as shown in Table \ref{TABLE:1}. 
It is observed that the correlation coefficients between prosodic phrase breaks for any two different emotions are less than 1, while the same emotion paris are always 1. 
This suggests that the prosodic phrase-breaking patterns vary across emotions, indicating that prosodic phrase breaks may be emotion-specific. 
To this end, we will study the emotion-aware prosodic phrasing scheme to achieve expressive TTS.

\section{E{\MakeLowercase{mo}}PP: Methodology}
\label{sec:model}
The study on prosodic pause prediction with deep neural networks has achieved laudable results. However, such prosodic phrase breaks are typically linguistically instead of emotionally motivated. In this work, we introduce a novel approach that first extracts the speaker's emotional state from the text. This emotion information is combined with the text and used as an input for predicting prosodic pauses that align with the emotional context of the text.

\subsection{Overall Architecture}
\begin{figure}[!th]
\centering
\setlength{\abovecaptionskip}{-0mm}   
\centerline{\includegraphics[width=1\linewidth]{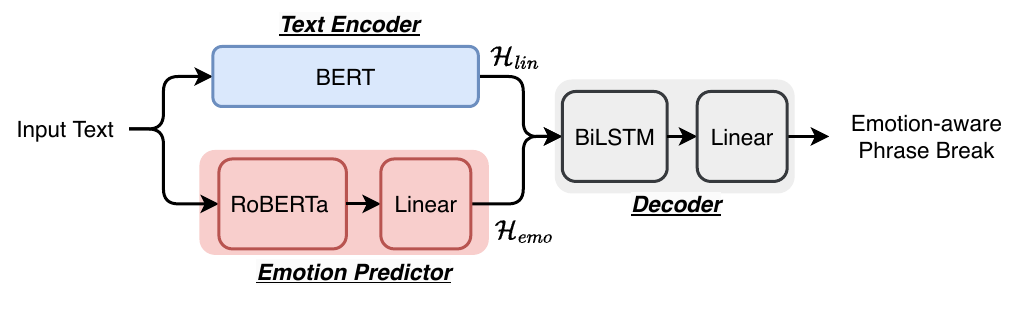}}
\vspace{-5mm}
\caption{The overall architecture of EmoPP, consists of text encoder, emotion predictor and decoder.}
\label{fig:fig1}
\vspace{-5mm}
\end{figure}

The overall architecture of our model is shown in Fig \ref{fig:fig1}. Our EmoPP consists of Text Encoder, Emotion Predictor, and Decoder. The text encoder aims to extract the linguistic feature from the input text. Emotion predictor seeks to infer the emotional category of the input text. The decoder takes both the linguistic feature and the emotional cues as input to predict the emotion-aware phrase breaks.

\subsubsection{Text Encoder}
Let's use $X$ to denote the input text. In view of the powerful semantic modeling of BERT \cite{devlin2018bert}, we adopt BERT as our text encoder to extract the word-level linguistic feature $\mathcal{H}_{lin}$ of the input text.
\begin{equation}
    \mathcal{H}_{lin} = Enc_{text}(X)
     \vspace{-3mm}
\end{equation}

\subsubsection{Emotion Predictor}
The emotion predictor consists of a RoBERTa layer and a linear layer. 
Note that RoBERTa is a variant of BERT with new training strategies \cite{liu2019roberta}, such as removing the \textit{next sentence prediction} objective and training on longer sequences. Owing to the exceptional performance of RoBERTa in various emotion classification tasks \cite{adoma2020comparative}, the RoBERTa layer is added to infer the emotional cues from the input text. After that, the linear layer is used to predict the final emotion category label. The predicted emotion labels are then embedded into emotion embedding $\mathcal{H}_{emo}$ through an embedding layer. 
\begin{equation}
    \mathcal{H}_{emo} = Pre_{emotion}(X)
\end{equation}

At last, the emotion embedding $\mathcal{H}_{emo}$ is concatenated with the linguistic feature $\mathcal{H}_{lin}$ of the text encoder to form a joint embedding $\mathcal{H}$, which will be fed into the Decoder.
\begin{equation}
    \mathcal{H} = concat (\mathcal{H}_{lin}, \mathcal{H}_{emo}) 
\end{equation}
where $concat(\cdot)$ means concatenate the $\mathcal{H}_{lin}$ with the $\mathcal{H}_{emo}$ along the last dimension.

 \vspace{-3mm}
\subsubsection{Decoder}
The decoder consists of a BiLSTM layer and a linear layer. The BiLSTM reads $\mathcal{H}$ to summarize long-term time dependencies and sequential characteristics into a representative feature.  
To prevent overfitting, dropout layers are applied after the BiLSTM layer. The output from the BiLSTM layer is passed through a linear layer to generate logits that are used to predict whether a pause exists at each word within the text. Finally, we obtain the final phrase break sequence $Y$. 
\begin{equation}
    Y = Dec (\mathcal{H}) 
\end{equation}
 
\subsection{Loss Functions}
 
It's worth mentioning that the emotion predictor is jointly trained with the whole network, which allows the EmoPP to perform both emotion prediction and prosodic pause prediction. Therefore, the total loss function include two parts $\mathcal{L}_{emo}$ and $\mathcal{L}_{pp}$. $\mathcal{L}_{emo}$ aims to make the output of the emotion predictor close to the true emotion category of the corresponding speech of that text, and the $\mathcal{L}_{pp}$ is used to make the output of the decoder close to the true sequence of prosodic phrase break sequence.
\begin{equation}
    \mathcal{L} = \mathcal{L}_{emo} + \alpha * \mathcal{L}_{pp}
   \label{eq1}
\end{equation}
where $\alpha$ is the balance factor between two loss functions.

In this way, we can leverage emotional information to aid in prosodic break prediction, thereby improving the model's ability to generate the prosodic break sequence that is consistent with the emotional expression of the utterance.

\section{Experiments and Results}
\label{sec:exp}
We validate the EmoPP with the IEMOCAP \cite{busso2008iemocap} dataset. 
To ensure class balance, we select data corresponding to five commonly observed emotions: neutral, happy, angry, sad, and surprised. 
Moreover, following the automatic break extraction pipeline, as introduced in Section \ref{sec:dataobs}, we derive the phrase break sequence for all utterances of IEMOCAP as the phrase break prediction training data.
 
\subsection{Experimental Setup}

For the baseline model, The dimension of word embeddings was 300. The hidden size and projection size for each BiLSTM layer were both 512. For the proposed model, we configured the BERT as bert-base-uncased1 \footnote{https://huggingface.co/bert-base-uncased/blob/main/config.json} and the RoBERTa as roberta-base2 \footnote{https://huggingface.co/roberta-base/blob/main/config.json}. The dimensions of the hidden sequence and emotion embeddings were 768. During training, each mini-batch had 16 sentences. We used the Adam optimizer and set the initial value of the dynamic learning rate to 1 $\times 10^{-5}$. The balance factor in Eq. \ref{eq1} is set to 0.7 based on experience.
We set the number of training epochs to 10. The models that performed best on the validation set during these iterations were saved to make comparisons.
To prevent training instability or the occurrence of gradient explosion when the norm of gradients becomes excessively large, we employ gradient clipping and set the norm threshold of gradients to 10.
 
\subsection{Comparative Study}
We develop three phrase break prediction models for a comparative study, that include the 1) \textbf{BiLSTM} \cite{klimkov2018phrase}: the classical system that takes BiLSTM as the backbone; 2) \textbf{BERT + BiLSTM} \cite{vadapalli2023investigation}: the advanced system that adopts BERT to extract the linguistic feature; 3) \textbf{EmoPP (Ours)}: the proposed emotion-aware prosodic phrasing model and 4) \textbf{w/o RoBERTa}: the ablation system aims to validate the RoBERTa module of emotion predictor. We replace the RoBERTa with a simple linear layer.

\begin{table}[t]
\caption{Performance comparison of baseline models and EmoPP model on IEMOCAP dataset.
}
\label{Table:2}
\centering
\small
\begin{tabular}{l|ccc}
\bottomrule
    \textbf{Systems}    & \textbf{Precision}   & \textbf{Recall}   & \textbf{F1-Score}  \\ \hline
 BiLSTM         & 75.08                         & 73.08                         & 73.90      \\  
 BERT + BiLSTM  &  78.49                 & 76.73                         & 77.48       \\ \hline
\textbf{EmoPP (Ours)}                      & \textbf{78.95}                         & \textbf{77.95}                & \textbf{78.43} \\  \hline
\quad $w/o$ RoBERTa &  77.76  &  73.57  & 74.95  \\ \bottomrule
\hline
\end{tabular}
\end{table}

\begin{figure}[!th]
\centering
\setlength{\abovecaptionskip}{-0mm}   
\centerline{\includegraphics[width=1.1\linewidth]{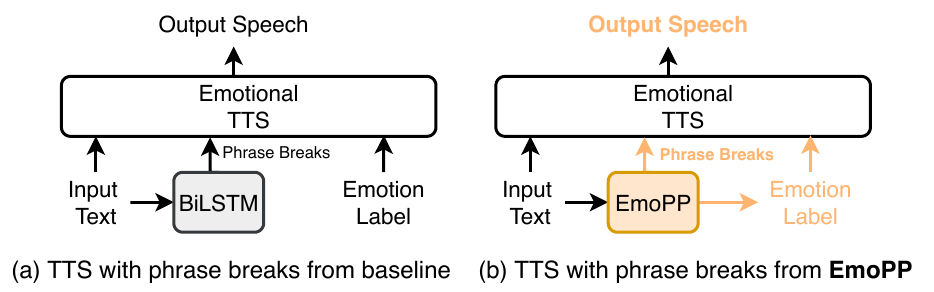}}
\vspace{-3mm}
\caption{Two TTS systems with different phrase breaks, predicted from BiLSTM baseline and the proposed EmoPP respectively.}
\vspace{-5mm}
\label{fig:fig2}
\end{figure}

\subsection{Objective Evaluations}
We report the objective results in terms of Precision (P), Recall (R) and F-score (F) which is defined as the harmonic mean of the P and R. F values range from 0 to 1, with a higher value indicating better performance.

We select 100 test samples from the test set randomly and report all results in Table \ref{Table:2}. We observe that EmoPP performs best in terms of phrase break prediction accuracy. Specifically, the Precision, Recall, and F1-Score achieve the optimal value among all systems. It suggests that our EmoPP incorporates the emotional cues and contributes to generating emotion-aware phrase breaks for the utterance. To demonstrate that our performance improvement is not due to the addition of a RoBERTa with a huge number of parameters, we can check the results of the last row of Table \ref{Table:2}. The ablation results show that the F1-Score performance dropped by a large margin. More importantly, although the F1-Score of $w/o$ RoBERTa is lower than that of EmoPP, its value is still higher than the BiLSTM baseline. We can find that the emotional cues of utterance still play a key role in prosodic phrasing, which is encouraging.

 \vspace{-2mm}
\subsection{Subjective Evaluations}
 \vspace{-2mm}
To further validate our EmoPP in terms of human perception, we build two emotional TTS systems that take both input text and the phrase breaks information as input. As shown in Fig. \ref{fig:fig2}, the phrase break information of (a) is obtained by the BiLSTM model, while (b) is obtained by our EmoPP. The emotional TTS is trained with an emotional conversational TTS dataset, DailyTalk \cite{lee2023dailytalk} \footnote{We attempted to train the emotional TTS model using the IEMOCAP dataset. However, the synthesized speech produced significant noise. Since IEMOCAP was not originally designed for TTS purposes, it is not optimal for our subjective test.}, and implemented by this project \footnote{\url{https://github.com/keonlee9420/DailyTalk}}.
We invite 10 volunteers and each volunteer is asked to listen to 100 samples to rate the prosody expressiveness with 5-scale Expressive Mean Opinion Score (EMOS). Note that EMOS just focuses on the emotion expressiveness performance for all samples.

The results are reported in Table \ref{Table:3}. We observe that the EMOS score of ``TTS with EmoPP'' system is higher than the ``TTS with BiLSTM'', indicating that the emotion-aware prosodic phrasing indeed contributes to the emotion expressive rendering. The subjective evaluation further supports the effectiveness of our EmoPP in terms of expressive modeling of TTS.

\begin{table}[t]
\caption{EMOS results of two TTS systems. 
}
\label{Table:3}
\centering
\small
\begin{tabular}{p{3cm}<{\centering}|p{2.3cm}<{\centering}}
\bottomrule
     \textbf{Systems}   & \textbf{EMOS}    \\ \hline
 TTS with BiLSTM         & 3.84 $\pm$ 0.09                  \\  
 TTS with EmoPP  & 4.09 $\pm$ 0.05        \\  
  \bottomrule\hline
\end{tabular}
\end{table}


\begin{figure}[!t]
\centering
\setlength{\abovecaptionskip}{-0mm}   
\begin{minipage}{0.7\linewidth}
  \centerline{
  \includegraphics[width= \linewidth]{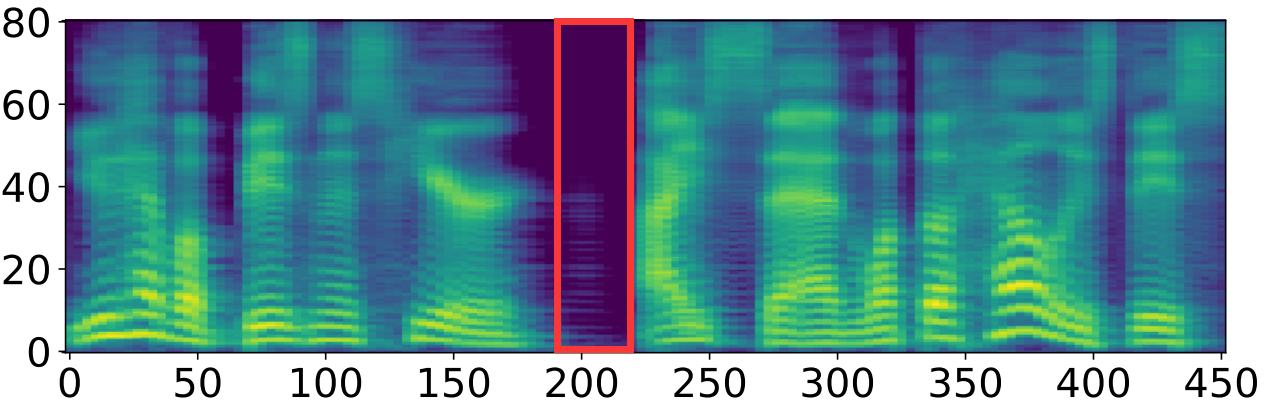}
  }
  \vspace{-0.5mm}
  \centerline{(a) {\small TTS with BiLSTM} }
\end{minipage}
\vfill
\begin{minipage}{0.7\linewidth}
  \centerline{
  \includegraphics[width= \linewidth]{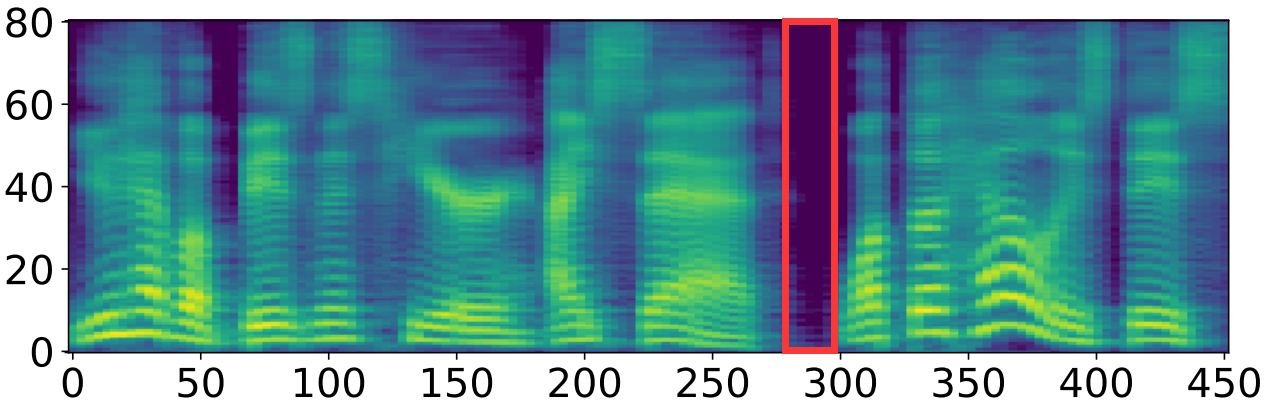}
  }
  \vspace{-0.5mm}
  \centerline{(b) {\small TTS with EmoPP} }
\end{minipage}
\vspace{-2.6mm}
\caption{Visualizations of the generated mel-spectrograms by two TTS systems. The red box indicates the phrase breaks.}
\label{fig:fig4}
\vspace{-4mm}
\end{figure}
 
\subsection{Visualization Analysis}
In this section, we visualize the Mel-spectrum features to further validate the EmoPP intuitively. We select the synthesized speech for text ``You have a business here. I said what the hell is this?'' from TTS models (a) and (b) in Fig. \ref{fig:fig2}.

The plotted Mel-spectrum features are shown in Fig. \ref{fig:fig4}. We find that with the addition of the emotion-aware phrase break after the word ``said'', the ``angry'' emotion expressed in this utterance is more pronounced, enhancing the emotional expressiveness of the speech.

\section{Conclusion}
\label{sec:con}
In this paper, we proposed an emotion-aware prosodic phrasing scheme, termed EmoPP, for expressive modeling of TTS. The additional designed emotion predictor on the basis of text encoder and decoder allows the EmoPP to mine the emotional cues of utterance accurately and predict appropriate phrase breaks. The objective and subjective experimental results suggest that our EmoPP outperforms all baselines in terms of break prediction accuracy and prosody expressive rendering for emotional TTS. In future work, we will further improve the model architecture and validate it on more datasets.

\bibliographystyle{IEEEbib}

{\footnotesize
\bibliography{strings}}

\begin{thebibliography}{10}

\bibitem{vadapalli2016investigation}
Anandaswarup Vadapalli and Suryakanth~V Gangashetty,
\newblock ``An investigation of recurrent neural network architectures using
  word embeddings for phrase break prediction.,''
\newblock in {\em Interspeech}, 2016, pp. 2308--2312.

\bibitem{li2019neural}
Naihan Li, Shujie Liu, Yanqing Liu, Sheng Zhao, and Ming Liu,
\newblock ``Neural speech synthesis with transformer network,''
\newblock in {\em Proceedings of the AAAI conference on artificial
  intelligence}, 2019, vol.~33, pp. 6706--6713.

\bibitem{shen2018natural}
Jonathan Shen, Ruoming Pang, Ron~J Weiss, Mike Schuster, Navdeep Jaitly,
  Zongheng Yang, Zhifeng Chen, Yu~Zhang, Yuxuan Wang, Rj~Skerrv-Ryan, et~al.,
\newblock ``Natural tts synthesis by conditioning wavenet on mel spectrogram
  predictions,''
\newblock in {\em 2018 IEEE international conference on acoustics, speech and
  signal processing (ICASSP)}. IEEE, 2018, pp. 4779--4783.

\bibitem{ping2017deep}
Wei Ping, Kainan Peng, Andrew Gibiansky, Sercan~O Arik, Ajay Kannan, Sharan
  Narang, Jonathan Raiman, and John Miller,
\newblock ``Deep voice 3: Scaling text-to-speech with convolutional sequence
  learning,''
\newblock {\em arXiv preprint arXiv:1710.07654}, 2017.

\bibitem{liu2020exploiting}
Rui Liu, Berrak Sisman, Feilong Bao, Jichen Yang, Guanglai Gao, and Haizhou Li,
\newblock ``Exploiting morphological and phonological features to improve
  prosodic phrasing for mongolian speech synthesis,''
\newblock {\em IEEE/ACM Transactions on Audio, Speech, and Language
  Processing}, vol. 29, pp. 274--285, 2020.

\bibitem{frazier2006prosodic}
Lyn Frazier, Katy Carlson, and Charles Clifton,
\newblock ``Prosodic phrasing is central to language comprehension,''
\newblock {\em Trends in cognitive sciences}, vol. 10, no. 6, pp. 244--249,
  2006.

\bibitem{keri2007pause}
Venkatesh Keri, Sathish~Chandra Pammi, and Kishore Prahallad,
\newblock ``Pause prediction from lexical and syntax information,''
\newblock in {\em Proceedings of International Conference on Natural Language
  Processing (ICON)}, 2007.

\bibitem{klimkov2018phrase}
Viacheslav Klimkov, Adam Nadolski, Alexis Moinet, Bartosz Putrycz, Roberto
  Barra-Chicote, Tom Merritt, and Thomas Drugman,
\newblock ``Phrase break prediction for long-form reading tts: Exploiting text
  structure information,''
\newblock 2018.

\bibitem{watts2011unsupervised}
Oliver Watts, Junichi Yamagishi, and Simon King,
\newblock ``Unsupervised continuous-valued word features for phrase-break
  prediction without a part-of-speech tagger.,''
\newblock in {\em Interspeech 2011-12th annual Conference of the International
  Speech Communication Association}. ISCA, 2011, pp. 2157--2160.

\bibitem{si2022modelling}
Chen Si, Caicai Zhang, Puiyin Lau, Yike Yang, and Bei Li,
\newblock ``Modelling representations in speech normalization of prosodic
  cues,''
\newblock {\em Scientific Reports}, vol. 12, no. 1, pp. 14635, 2022.

\bibitem{yi2023adversarial}
Jiangyan Yi, Jianhua Tao, Ruibo Fu, Tao Wang, Chu~Yuan Zhang, and Chenglong
  Wang,
\newblock ``Adversarial multi-task learning for mandarin prosodic boundary
  prediction with multi-modal embeddings,''
\newblock {\em IEEE/ACM Transactions on Audio, Speech, and Language
  Processing}, vol. 31, pp. 2963--2973, 2023.

\bibitem{qian2010automatic}
Yao Qian, Zhizheng Wu, Xuezhe Ma, and Frank Soong,
\newblock ``Automatic prosody prediction and detection with conditional random
  field (crf) models,''
\newblock in {\em 2010 7th International Symposium on Chinese Spoken Language
  Processing}. IEEE, 2010, pp. 135--138.

\bibitem{liu2018mongolian}
Rui Liu, Feilong Bao, Guanglai Gao, and Yonghe Wang,
\newblock ``Mongolian text-to-speech system based on deep neural network,''
\newblock in {\em Man-Machine Speech Communication: 14th National Conference,
  NCMMSC 2017, Lianyungang, China, October 11--13, 2017, Revised Selected
  Papers 14}. Springer, 2018, pp. 99--108.

\bibitem{chen1998rnn}
Sin-Horng Chen, Shaw-Hwa Hwang, and Yih-Ru Wang,
\newblock ``An rnn-based prosodic information synthesizer for mandarin
  text-to-speech,''
\newblock {\em IEEE transactions on speech and audio processing}, vol. 6, no.
  3, pp. 226--239, 1998.

\bibitem{liu2018improving}
Rui Liu, Feilong Bao, Guanglai Gao, Hui Zhang, and Yonghe Wang,
\newblock ``Improving mongolian phrase break prediction by using syllable and
  morphological embeddings with bilstm model.,''
\newblock in {\em Interspeech}, 2018, pp. 57--61.

\bibitem{du2019prosodic}
Yao Du, Zhiyong Wu, Shiyin Kang, Dan Su, Dong Yu, and Helen Meng,
\newblock ``Prosodic structure prediction using deep self-attention neural
  network,''
\newblock in {\em 2019 Asia-Pacific Signal and Information Processing
  Association Annual Summit and Conference (APSIPA ASC)}. IEEE, 2019, pp.
  320--324.

\bibitem{duez1982silent}
Danielle Duez,
\newblock ``Silent and non-silent pauses in three speech styles,''
\newblock {\em Language and speech}, vol. 25, no. 1, pp. 11--28, 1982.

\bibitem{zhou2022emotional}
Kun Zhou, Berrak Sisman, Rui Liu, and Haizhou Li,
\newblock ``Emotional voice conversion: Theory, databases and esd,''
\newblock {\em Speech Communication}, vol. 137, pp. 1--18, 2022.

\bibitem{yang2023duration}
Dong Yang, Tomoki Koriyama, Yuki Saito, Takaaki Saeki, Detai Xin, and Hiroshi
  Saruwatari,
\newblock ``Duration-aware pause insertion using pre-trained language model for
  multi-speaker text-to-speech,''
\newblock in {\em ICASSP 2023-2023 IEEE International Conference on Acoustics,
  Speech and Signal Processing (ICASSP)}. IEEE, 2023, pp. 1--5.

\bibitem{smc}
``Simple matching coefficient,''
  \url{https://people.revoledu.com/kardi/tutorial/Similarity/SimpleMatching.html}.

\bibitem{louw2016speaker}
Johannes~A Louw and Avashlin Moodley,
\newblock ``Speaker specific phrase break modeling with conditional random
  fields for text-to-speech,''
\newblock in {\em 2016 Pattern Recognition Association of South Africa and
  Robotics and Mechatronics International Conference (PRASA-RobMech)}. IEEE,
  2016, pp. 1--6.

\bibitem{prahallad2010learning}
Kishore Prahallad, E~Veera Raghavendra, and Alan~W Black,
\newblock ``Learning speaker-specific phrase breaks for text-to-speech
  systems,''
\newblock in {\em Seventh ISCA Workshop on Speech Synthesis}, 2010.

\bibitem{devlin2018bert}
Jacob Devlin, Ming-Wei Chang, Kenton Lee, and Kristina Toutanova,
\newblock ``Bert: Pre-training of deep bidirectional transformers for language
  understanding,''
\newblock {\em arXiv preprint arXiv:1810.04805}, 2018.

\bibitem{liu2019roberta}
Yinhan Liu, Myle Ott, Naman Goyal, Jingfei Du, Mandar Joshi, Danqi Chen, Omer
  Levy, Mike Lewis, Luke Zettlemoyer, and Veselin Stoyanov,
\newblock ``Roberta: A robustly optimized bert pretraining approach,''
\newblock {\em arXiv preprint arXiv:1907.11692}, 2019.

\bibitem{adoma2020comparative}
Acheampong~Francisca Adoma, Nunoo-Mensah Henry, and Wenyu Chen,
\newblock ``Comparative analyses of bert, roberta, distilbert, and xlnet for
  text-based emotion recognition,''
\newblock in {\em 2020 17th International Computer Conference on Wavelet Active
  Media Technology and Information Processing (ICCWAMTIP)}. IEEE, 2020, pp.
  117--121.

\bibitem{busso2008iemocap}
Carlos Busso, Murtaza Bulut, Chi-Chun Lee, Abe Kazemzadeh, Emily Mower, Samuel
  Kim, Jeannette~N Chang, Sungbok Lee, and Shrikanth~S Narayanan,
\newblock ``Iemocap: Interactive emotional dyadic motion capture database,''
\newblock {\em Language resources and evaluation}, vol. 42, pp. 335--359, 2008.

\bibitem{vadapalli2023investigation}
Anandaswarup Vadapalli,
\newblock ``An investigation of speaker independent phrase break models in
  end-to-end tts systems,''
\newblock {\em arXiv preprint arXiv:2304.04157}, 2023.

\bibitem{lee2023dailytalk}
Keon Lee, Kyumin Park, and Daeyoung Kim,
\newblock ``Dailytalk: Spoken dialogue dataset for conversational
  text-to-speech,''
\newblock in {\em ICASSP 2023-2023 IEEE International Conference on Acoustics,
  Speech and Signal Processing (ICASSP)}. IEEE, 2023, pp. 1--5.

\end{thebibliography}

\end{document}